\documentclass{article}
\usepackage{ijcai22}

\usepackage{times}
\usepackage{soul}
\usepackage{url}
\usepackage[hidelinks]{hyperref}
\usepackage[utf8]{inputenc}
\usepackage[small]{caption}
\usepackage{graphicx}
\usepackage{amsmath}
\usepackage{amsthm}
\usepackage{booktabs}
\usepackage{algorithm}
\usepackage{algorithmic}
\title{Learning to Reverse DNNs from AI Programs Automatically}

\author{
Simin Chen$^1$
\and
Hamed Khanpour$^2$\and
Cong Liu$^{1}$\And
Wei Yang$^1$
\affiliations
$^1$The University of Texas at Dallas\\
$^2$Microsoft Research
\emails
\{Simin.Chen, Cong, Wei.Yang\}@utdallas.com,
Hamed.Khanpour@microsoft.com,
}

\usepackage{booktabs}
\usepackage{multirow}
\usepackage{xcolor}
\usepackage{adjustbox}
\usepackage{bbding}
\usepackage{xspace}
\usepackage{ dsfont }
\usepackage{amsmath}

\newcommand{\eg}{\hbox{\emph{e.g.}}\xspace}
\newcommand{\ie}{\hbox{\emph{i.e.}}\xspace}

\newcommand{\etc}{\hbox{\emph{etc.}}\xspace}

\newcommand\figref[1]{Fig.~\ref{#1}}

\newcommand\tabref[1]{Table~\ref{#1}}

\newcommand\equref[1]{Eq.(\ref{#1})}

\usepackage{ amssymb }

\ifodd 1

\else

\fi

\def \tool{\texttt{NNReverse}\xspace}

\begin{document}

\maketitle
\begin{abstract}

    With the privatization deployment of DNNs on edge devices, the security of on-device DNNs has raised significant concern.    
    To quantify the model leakage risk of on-device DNNs automatically,  we propose \tool, the first learning-based method which can reverse DNNs from AI programs without domain knowledge.
    \tool trains a representation model to represent the semantics of binary code for DNN layers.  By searching the most similar function in our database, \tool infers the layer type of a given function's binary code.
    To represent assembly instructions semantics precisely, \tool proposes a more fine-grained embedding model to represent the textual and structural-semantic of assembly functions.

\end{abstract}

\section{Introduction}

Over the last few years, deploying Deep Neural Networks~(DNNs) on edge devices is becoming a popular trend for giant AI providers.
The AI providers compile the private high-quality DNN models into stand-alone programs~\cite{li2020deep,chen2018tvm,rotem2018glow,cyphers2018intel} and would like to sell them to other companies, organizations, and governments with a license fee.
This edge-deployment situation further exacerbates the risk of model leakage. 
The DNN model leakage will result in the following three consequences.
First, the potential loss of intellectual property. DNNs parameters are typically trained from TB datasets with high training costs. For example, training a DNN on the google cloud TPU will cost more than \$400K~\cite{devlin2018bert,raffel2019exploring,zhu2021hermes}. Thus, DNNs architecture and parameters are precious intellectual property of the developers. 
Second, the potential infringement of privacy. DNNs parameters have been proven to remember sensitive information~(\eg SSN) in the DNNs training data~\cite{carlini2020extracting}. Therefore, the leakage of DNN parameters will result in the privacy leakage of training data.
Last but not least, the potential risk of being attacked by adversaries. DNNs are vulnerable to adversarial attacks~\cite{carlini2017towards}. If the attackers obtain the parameters, they could easily compute DNNs gradients to launch white-box attacks.
For Example, reversing Apple's image hash model, NeuralHash~\footnote{\href{https://www.schneier.com/blog/archives/2021/08/apples-neuralhash-algorithm-has-been-reverse-engineered.html}{Apples-neuralhash-algorithm-has-been-reverse-engineered}}, will enable  attackers to break the content protection provided by macOS.
Therefore, conducting penetration testing to protect DNNs' privacy is essential.

\begin{figure}
    \centering
    \includegraphics[width=0.46\textwidth]{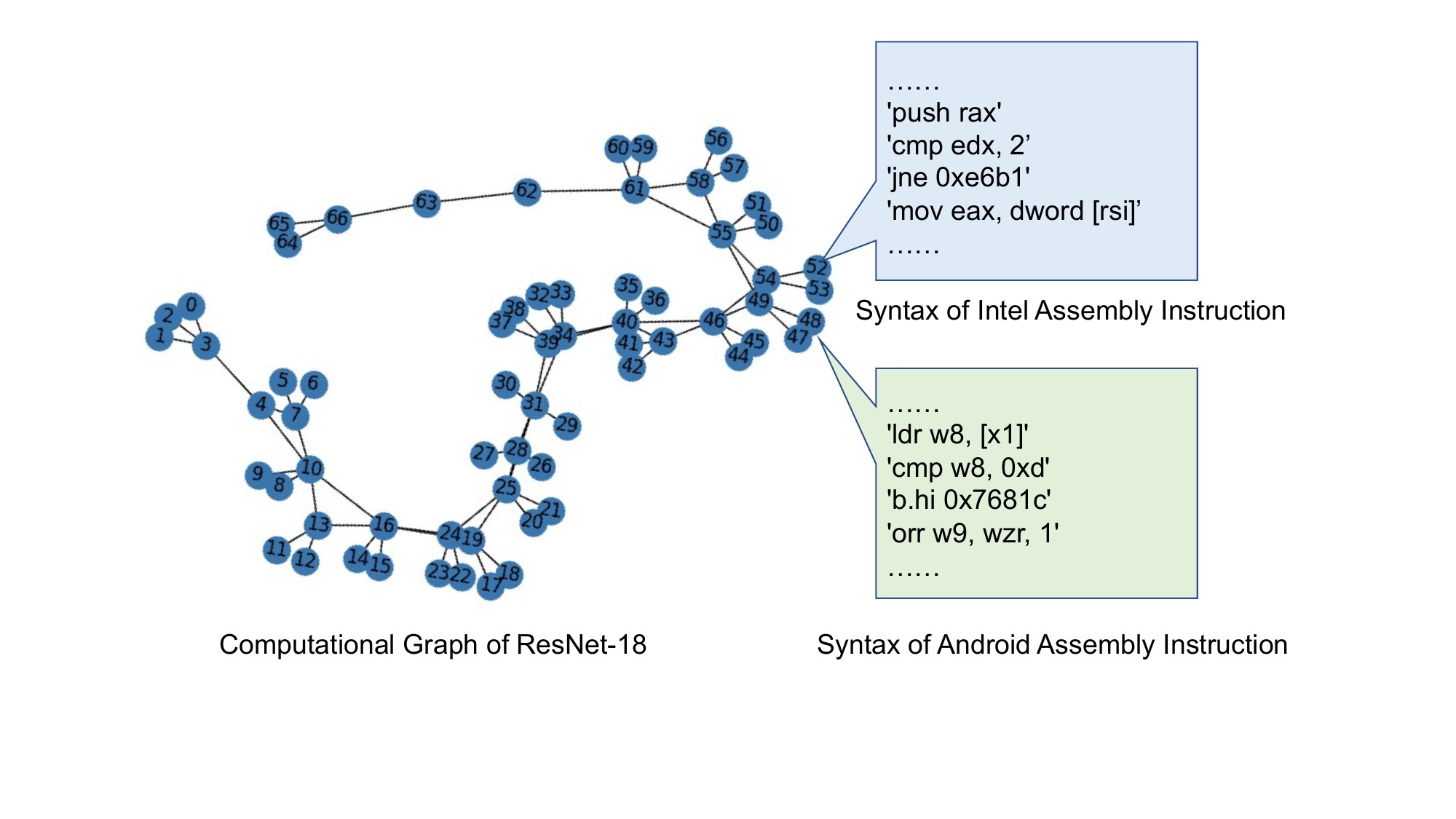}
    \caption{Assembly instructions of ResNet-18 on different platforms}
    \label{fig:challenge}
\end{figure}

Quantifying  model leakage risk of on-device DNNs automatically  is a challenging task because of the following two reasons: 
(1) Reversing DNN architectures requires understanding the semantics of the machine code of each DNN layer. 
Machine code is incomprehensible to humans; unfortunately, modern DNN usually contains hundreds of layers, and each layer corresponds to a function that needs to infer the semantic. 
(2) The cross-platform deployment of DNN models increase the difficulty of understanding the semantics of the code. DNNs may be deployed on different hardware platforms. The machine code on different hardware platforms have different syntax. Therefore, understanding the semantics of machine code on different hardware platforms requires different domain-specific knowledge.
\figref{fig:challenge} shows the challenge in reversing DNNs, the left figure shows the computational graph of Resnet-18, reversing ResNet-18 requires inferring the semantic of each node in the graph, and the right figure shows different assembly instructions syntax on different hardware platforms.

In this paper, we propose \tool, the first static model stealing attack to reverse DNNs from AI programs automatically.
Different from existing model extraction attacks~\cite{jagielski2020high,oh2019towards,tramer2016stealing}, which approximate the victim DNNs functionality rather than reversing DNN architectures~\cite{krishna2019thieves}.
Our approach does not train a substitute model to approximate input-output pairs from the victim DNNs.
Instead, our approach learns and infer the semantic of machine code for different hardware platforms automatically.

Our intuition is that although DNN architectures are complicated, the basic types~(\ie \textit{conv}, \textit{dense}) of DNN layers are enumerable. 
Thus, it is possible to reverse DNN architectures by inferring each DNN layer type. Furthermore, because different DNN layers are implemented as different binary functions, we can infer DNN layer type by inferring semantic corresponding binary functions.
Specifically, our attack includes two phases: offline phase and online phase. 
In the offline phase, we build a database to collect the binary of each optimized kernel layer, then we train a semantic representation model on the decompiled functions.
The semantic representation model can be trained with machine code from different hardware platforms, thus, \tool could reverse DNNs for different hardware platforms.
Although some semantic representation approaches for NLP fields have been proposed, they are not suitable for assembly functions because they ignore the topology semantic in assembly functions. 
We propose a new embedding algorithm to overcome this limitation and represent assembly functions more precisely, which combines syntax semantic representation and topology semantic.
In the online phase, we decompile the victim DNN executable and collect the binaries. By the pre-collected knowledge in the database, we reconstruct the whole DNN model.
We summarize our contribution as follows:

\begin{itemize}
    \item \textbf{Characterizing}: We characterize the model leakage risk of on-device DNNs. Specifically, we show that on-device DNNs could be reversed easily.

    \item \textbf{Approach}: We propose a new fine-grained embedding approach for representing assembly functions, which combines syntax representation and topology structure representation.
    
    \item \textbf{Evaluation}: We implement \tool and evaluate \tool on two datasets. The evaluation results show \tool can achieve higher accuracy in searching similar semantic functions than baseline methods, and we apply \tool to reverse real-world DNNs as a case study. Results show that \tool can reverse the DNNs without accuracy loss.
    
    \item \textbf{Dataset}: We open-source our decompiled assembly function dataset~\footnote{\href{https://drive.google.com/drive/folders/1dsGyPqfwRa9JPuBxp1mGKODwTWFlDqrc?usp=share_link}{https://drive.google.com/file/d/NNReverse}}, including more than 80k functions compiled for four hardware platforms. Our dataset is the first dataset for DNN reverse engineering, contributing to future research on on-device DNN security.
    
\end{itemize}

\section{Background \& Related Work}

\subsection{DNN Deployment}

\begin{figure}[t!]
    \centering
    \includegraphics[width=0.45\textwidth]{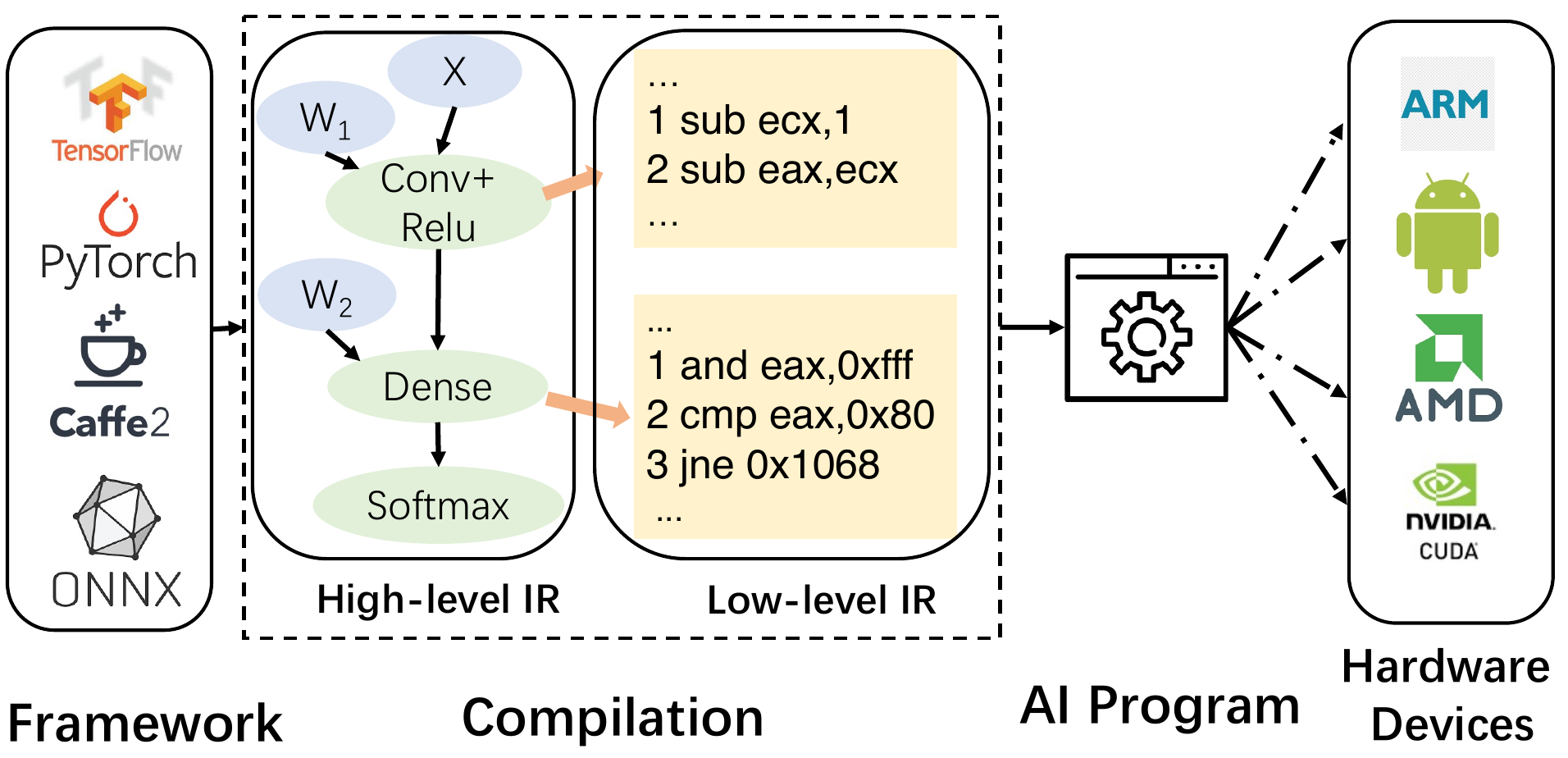}
    \caption{The process of compiling DNNs into programs.}
    \label{fig:deploy_process}
\end{figure}

DL compilers are used to generate portable AI programs from high-level DL frameworks such as \texttt{Pytorch}, \texttt{TensorFlow}, and \texttt{Caffe}.
The generated AI programs are advanced optimized~(\eg Layer Fusion) to match the target hardware devices.
The working process of DL compiler is shown in \figref{fig:deploy_process}.
DL compilers accept DL models from high-level DL frameworks as input and first translate DL models into high-level intermediate representation~(IR).
The high-level IR~(as known as graph IR) is a computational graph, where nodes represent the atomic DL operators (Convolution, Pooling, Convolution+ReLu, \etc) or tensors, and the edges represent the direction of dataflow.
Then DL compilers translate each atomic DL operator into low-level IR and generate the executable code by optimizing the computation and memory access for the target hardware devices.
DL compilers enable running DL models on different hardware devices such as X86, ARM, NVIDIA, and TPU.
For more details about DL compilers, we refer readers to the existing work~\cite{li2020deep,chen2018tvm,rotem2018glow,cyphers2018intel}.
In this paper, we consider TVM~\cite{tvm}, a widely-used DL compiler to deploy DNNs. Given a DNNs from high-level frameworks~(\eg \texttt{PyTorch}), TVM will output three files: graph file, parameter file, and binary file.
Notice that, These files are necessary for compiled AI programs to run. 
There are similar functionality files for other DL compilers. For example, Apple Core ML applies the file formats \textit{.espresso.net} and \textit{.espresso.weights} to store computational graphs and weights. 

\subsection{Model Extraction Attacks}

\begin{table}[tbp]
  \centering

  \begin{adjustbox}{width=.5\textwidth,center}
        \begin{tabular}{l|l|l|c|c|c}
    \toprule
    \multirow{2}[4]{*}{Work} & \multirow{2}[4]{*}{Information } & \multirow{2}[4]{*}{Method} & \multirow{2}[4]{*}{Platform-Independent} & \multicolumn{2}{c}{Results} \\
\cmidrule{5-6}          &       &       &       & Architecture & Parameters  \\
    \midrule
    1~\cite{yan2020cache}     & Catche & Search & N     & \textcolor{red}{×}    & \textcolor{green}{$\surd$}   \\ 
    \midrule
    2~\cite{hua2018reverse}     & Accelerator & Search,Infer & Y    & \textcolor{green}{$\surd$}    & \textcolor{red}{×}    \\
    \midrule
    3~\cite{duddu2018stealing}     & Timing & Search & N     & \textcolor{green}{$\surd$}     & \textcolor{red}{×}    \\
    \midrule
    4~\cite{roberts2019model}     & Noise Input & Infer & Y     & P     & P  \\
    \midrule
    5~\cite{jagielski2020high}     & Queries & Infer & Y     & \textcolor{red}{×}     & \textcolor{red}{×}   \\
    \midrule
    \textbf{Ours} & \textbf{Binary files} & \textbf{Infer} & \textbf{Y} & \textcolor{green}{$\surd$}  & \textcolor{green}{$\surd$} \\
    \bottomrule
    \end{tabular}%
  \end{adjustbox}
    \caption{Y stands for applicable for specific device, N stands for device-independent. $\surd$ stands for fully recover, P stands for partial recover, × means cannot recover.}
  \label{tab:related}%
\end{table}%

As shown in Table \ref{tab:related}, a number of attack algorithms~\cite{jagielski2020high,oh2019towards,tramer2016stealing,zhu2021hermes} have been proposed to extract DNNs. 
We summarize the existing attacks into two categories based on their extraction fidelity.
The first type of existing attack is \textit{functionality attack}, they treat the victim model as a label oracle and train another model to approximate the input-output pair from the victim model.
These attacks can not steal the DNN architecture and the DNN parameters, instead, they just clone the functionality of the victim model.
The second type of existing attack is \textit{steal attack}.
The goal of this attack is to reverse the actual DNN architecture or parameters.
For example,
\cite{yan2020cache} introduced a cache-based side-channel attack to steal DNN architectures,
\cite{roberts2019model} assume the model  architecture and the output of the softmax layer is known, and replicate the model parameters by feeding noise inputs.
Different from the existing \textit{steal attack}, we propose an infer-based \textit{steal attack} which requires no assumption about the environment in which the DNN model is running.

\subsection{Binary Code Analysis}
In this section, we introduce some binary analysis techniques for reverse engineering.
Traditional binary analysis techniques often extract hand-crafted features by domain experts to match similar functions and infer binary function semantics.
Recently, motivated by the representation learning in NLP fields, researchers propose to apply learning-based approaches for binary code analysis.
Specifically, researchers propose to learn a function representation that is supposed to encode the binary function syntax and semantics in low dimensional vectors, known as function embeddings.
For example, inspired by Doc2Vec, Asm2Vec proposes to model binary code as documents and design a novel learning algorithm to embed binary functions.
However, the learning architectures adopted in existing approaches are limited in modeling long-range dependencies~(based on Doc2Vec). 
The binary code for the DNN operators is long-range dependent because of the loop operations. For example, the four-level loops in the Conv2D layer.
Unlike existing approaches, our approach models both instruction and topology semantics of binary code, thus representing the semantics more precisely.

\section{NNReverse}

\subsection{Problem Formulation}

In this paper, we consider an AI model privatization deployment scenario, where the DNN model developers compile their private DNN models into stand-alone AI programs and sell the programs to third-party customers with subscription or perpetual licensing.
The considered scenario is realistic because many AI models have been deployed on mobile devices (\eg, Apple NeuralHash, Mobile Translation App, AIoT)and this assumption is widely used in existing work.
The end-users can physically access the AI programs, especially the machine code~(executable code) of the AI programs.
The goal of malicious attackers is to reverse the DNNs from the AI program.
Specifically, they can reverse the architecture of the DNNs and reconstruct the DNNs with high-level frameworks~(\ie Pytorch) to compute the DNNs gradients to launch evasion attacks.
Formally, our goal includes: \textit{(i)} inferring the kernel operator type of each binary function in the victim AI-programs; \textit{(ii)} reconstructing whole DNNs with high-level DL frameworks.
Reversing on-device DNNs have a severe real-world impact because DNNs parameters are precious intellectual property, and reverse engineering can bridge the gap between white-box and black-box attacks for DNNs~(\eg, bypass Apple NeuralHash). 
Furthermore, understanding the semantic intricacies of deployed binary code makes automatically reversing on-device DNNs difficult.

\subsection{Design Overview}

\begin{figure}[tp!]


    \centering
    \includegraphics[width=0.46\textwidth]{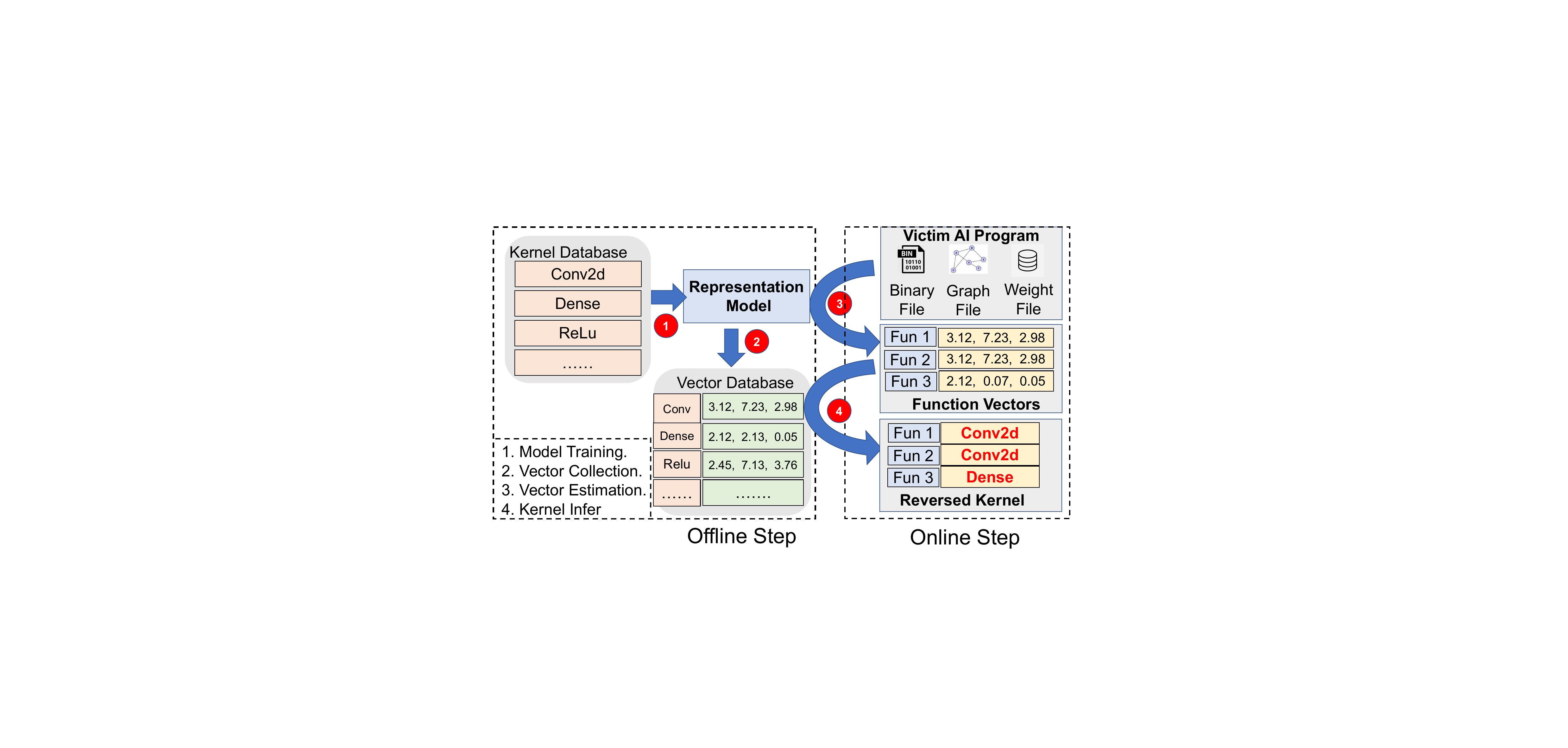}
    \caption{Design overview of our attack method}
    \label{fig:overview}
    

\end{figure}

As shown in \figref{fig:overview}, our approach includes offline and online steps.
In the offline step, we collect a set of DNNs with known architectures and use TVM~\cite{tvm}, a widely-used DL compiler,  to compile them into AI programs including a binary file for each program. We keep the debugging information for each binary file to keep a record of the kernel operator type of each binary function. We then train a semantic-representation model with these binary files.
In the online step, given a victim AI program, we first parse the AI program and decompile the parsed stripped binary file into a list of assembly functions.
Next, we feed each decompiled assembly function into the trained semantic representation model to estimate the vectors.
Finally, we compare the estimated vectors with the vectors in our offline database to infer the kernel operator type.
We could reverse the DNNs arch after inferring the kernel operator type and rebuild it with PyTorch by combing the weight files.

\subsection{Semantic Representation Model}
\label{sec:model}

\begin{figure*}[tp!]
    \centering
    \includegraphics[width=0.9\textwidth]{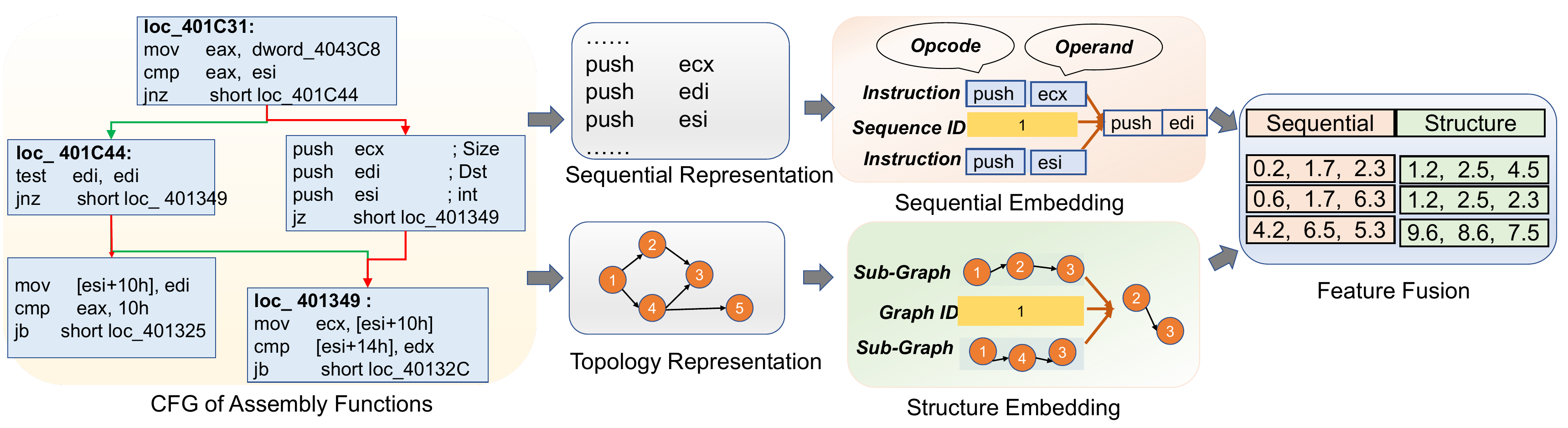}
    \caption{Design overview of our semantic representation model}
    \label{fig:approach}
\end{figure*}

In this section, we propose an embedding model to represent assembly functions as continuously distributed vectors.
As shown in \figref{fig:approach}, an assembly function can be represented as a control flow graph (CFG), where nodes are blocks consisting of sequential assembly instructions and edges are control flow paths.
Our embedding paradigm, in particular, fuses both text representation and structure representation from the assembly functions.

\paragraph{Text Embedding.}
We first model the assembly functions as sequential assembly instructions and seek to learn a function $\phi_{t}(\cdot)$, whose input is a sequential assembly instruction and output is a $\mathds{N}$ dimensional vector.
We embed the semantics of assembly instructions with more fine-grained modeling rather than treating each instruction as a token. Specifically, for each assembly instruction, we differentiate the \textit{opcode} and \textit{operands} in the instruction. Then we concatenate the vector of \textit{opcode} and \textit{operands} as the representation of the instruction. For instruction with more than one \textit{operands}, we use the average vector instead.
It can be formulated as:
\begin{equation}
    \mathcal{I}(ins) = \mathds{E}(\Omega(ins)) \; || \; \frac{1}{|\Theta(ins)| }\sum_{i=0}^{|\Theta(ins)|}\mathds{E}(\Theta(ins)_i)
\end{equation}
Where $ins$ represents an assembly instruction, $\Omega(\cdot)$ and $\Theta(\cdot)$ are the functions to parse \textit{opcode} and \textit{operands} respectively, and $\mathds{E}(\cdot)$ is the basic embedding layer that transfer tokens to corresponding numeric vectors.
Take the assembly instruction $ins = push, rbx$ as example, then $\Omega(ins) = push$ and $\Theta(ins) = rbx$.
Because the \textit{operands} may be a memory address~(\eg 0x73ff) or a immediate value~(\eg 0x16, 0x1024).
To eliminate the issue of out of vocabulary~(OOV), we propose a blur strategy.
Specifically, for memory address value and immediate value, we compare the value with 1000. 
If the value is less than 1000, we keep the original value, otherwise, we use token \textit{large} to replace the original value.

\begin{equation}
\label{eq:text}
    \sum_{j = 1}^{M}\sum_{i=1}^{|T(f_{j})|}P(\mathcal{I}_i(ins) \; | \; C_i, \phi_{t}(f_j) )
\end{equation}
After modeling assembly instruction, we seek to represent the semantic of the assembly function.
Specifically, we treat each instruction $\mathcal{I}(ins)$ as a ``word'' and apply a similar idea with doc2vec.
Formally, our training objective can be formulated as Equation \ref{eq:text}.
Where $M$ is the number of the training assembly functions, $f_j$ is the $j^{th}$ assembly function, $T(\cdot)$ is the text representation, $C_i$ is the neighbor instruction of  $\mathcal{I}_{i}(ins)$ and $\phi_{t}$ is the text embedding under learned.

\begin{table*}[tp]
  \centering

  \resizebox{0.75\textwidth}{!}{
    \begin{tabular}{c|c|ccc|ccc|ccc}
    \toprule
    \multicolumn{2}{c|}{\multirow{2}[2]{*}{\textbf{Dataset}}} & \multicolumn{3}{c|}{\textbf{P}} & \multicolumn{3}{c|}{\textbf{R}} & \multicolumn{3}{c}{\textbf{F1}} \\
    \multicolumn{2}{c|}{} & \textbf{Text} & \textbf{Structure} & \textbf{Combine} & \textbf{Text} & \textbf{Structure} & \textbf{Combine} & \textbf{Text} & \textbf{Structure} & \textbf{Combine} \\
    \midrule
          & \textbf{ARM} & 0.8595  & 0.5018  & \textbf{0.8663 } & \textbf{0.8875 } & 0.6665  & 0.8855  & 0.8733  & 0.5725  & \textbf{0.8758 } \\
          & \textbf{Aarch64} & 0.7562  & 0.3468  & \textbf{0.7688 } & \textbf{0.8406 } & 0.5388  & 0.8306  & 0.7962  & 0.4220  & \textbf{0.7985 } \\
    \textbf{Vision} & \textbf{Android} & \textbf{0.8419 } & 0.4578  & 0.8398  & 0.8741  & 0.6746  & \textbf{0.8766 } & 0.8577  & 0.5455  & \textbf{0.8578 } \\
          & \textbf{Intel} & 0.8427  & 0.4954  & \textbf{0.8444 } & \textbf{0.8725 } & 0.6719  & 0.8713  & 0.8574  & 0.5703  & \textbf{0.8576 } \\
          & \textbf{AVG} & 0.8251  & 0.4505  & \textbf{0.8298 } & \textbf{0.8687 } & 0.6380  & 0.8660  & 0.8462  & 0.5276  & \textbf{0.8474 } \\
    \midrule
          & \textbf{ARM} & 0.7232  & 0.2337  & \textbf{0.7610 } & 0.6859  & 0.3990  & \textbf{0.7104 } & 0.7041  & 0.2948  & \textbf{0.7348 } \\
          & \textbf{Aarch64} & 0.7301  & 0.5148  & \textbf{0.7342 } & 0.6224  & 0.2566  & \textbf{0.6519 } & 0.6720  & 0.3425  & \textbf{0.6906 } \\
    \textbf{Textual} & \textbf{Android} & \textbf{0.6836 } & 0.2007  & 0.6660  & 0.6872  & 0.2847  & \textbf{0.6882 } & \textbf{0.6854 } & 0.2354  & 0.6769  \\
          & \textbf{Intel} & \textbf{0.7557 } & 0.4085  & 0.7146  & 0.6889  & 0.3825  & \textbf{0.7050 } & \textbf{0.7208 } & 0.3951  & 0.7098  \\
          & \textbf{AVG} & \textbf{0.7232 } & 0.3394  & 0.7190  & 0.6711  & 0.3307  & \textbf{0.6889 } & 0.6956  & 0.3170  & \textbf{0.7030 } \\
    \bottomrule
    \end{tabular}%
}
  \caption{The results of proposed text embedding, structure embedding and combined embedding}
  \label{tab:combine}%
\end{table*}%

\begin{algorithm}[ht!]
	\renewcommand{\algorithmicrequire}{\textbf{Input:}}
	\renewcommand{\algorithmicensure}{\textbf{Output:}}
	\caption{Structure Embedding Algorithm}
	\label{alg:structure}
	\begin{algorithmic}[1]
	\REQUIRE G. \COMMENT{A set of CFGs topology structures}
	\REQUIRE D. \COMMENT{Maximum degree of rooted subgraphs}
	\STATE Initialization: Sample $\phi_s$ from $R^{G \times N}$
    	\FOR{$g_i \in G$}
        	\FOR{$n \in$ NODE$(g_i)$}
        	     	\FOR{$d = 0$ to $D$}
        	     	\STATE $sub_g$ = ComputeWLSubgraph($n, g_i, d$)
        	     	\STATE $\phi_s = \phi_s - \frac{\partial  -log\text{P} (sub_g \; | \; \phi_s(g_i))}{\partial \phi_s}$
        	        \ENDFOR
        	\ENDFOR
    	\ENDFOR

	\end{algorithmic}  
\end{algorithm}

\paragraph{Structure Embedding.}
Given a set of CFG topology structures, we intend to learn a distributed representation for every CFGs' topology structure.
Formally, we represent each CFG as a graph $G=(N, E)$, where $N$ are assembly blocks and $E$ are  connections between blocks. 
Our goal is to learn a embedding function $\phi_s(\cdot)$, whose input is $g \in G $ and output is a $\mathds{N}$ dimensional vector.
Our intuition is that two graphs are more topology similar if these two graphs have more topology similar sub-graphs.
Different from the semantic embedding that we apply random-walk paths as representation, applying sub-graphs will keep higher-order substructure and non-linear substructure information.
As shown in Algorithm \ref{alg:structure}.  Our algorithm require a set of graphs $G$ as the training corpus.
In each training iteration, we compute the WeisfeilerLehman sub-graphs~\cite{shervashidze2011weisfeiler} of each node in each graph $g_i$, then we try to maximize the possibility that the sub-graphs appear under the whole graph and update our structure embedding function $\phi_s(\cdot)$.
We repeat this process iteratively until the maximum epochs is reached.
 \begin{equation}
\label{eq:combine}
    \phi_{binary}(f) = \frac{\phi_{t}(T(f))}{| \phi_{t}(T(f))|} \; || \; \frac{\phi_{s}(S(f))}{|\phi_{s}(S(f))|}  
\end{equation}
\paragraph{Feature Fusion.}
 Finally, we combine the vectors from text embedding and structure embedding to obtain the final function embedding.
 Equation \ref{eq:combine} shows our fusion process, where $f$ is the assembly function, $T(f)$ and $S(f)$ are the text representation and the structure representation of $f$ respectively,  $\phi_{t}$ and $\phi_{s}$ are the text embedding function and structure embedding function, and $||$ is the vector concatenation.
\begin{equation}
\label{eq:cosine}
    kernel(\mathcal{B}) = name_{i} \quad i = \text{argmax}_{j} \; \frac{v \cdot u_j}{||v||\times||u_j||}  
\end{equation}
\paragraph{Binary Matching.}
After the model is well trained, we obtain the pre-trained database $U = \{(u_j, name_j) \}$, where $u_j$ and $name_j$  are the vector and the kernel type  of the $j^{th}$ function in the database. During the inference step, given a binary function $\mathcal{B}$, we first feed $\mathcal{B}$ into our model to get a vector $v$. we then apply \equref{eq:cosine} to assign $\mathcal{B}$ a known kernel type.

\begin{table*}[t]
  \centering
  \resizebox{0.95\textwidth}{!}{
    \begin{tabular}{c|c|ccc|ccc|ccc|ccc|ccc|ccc}
    \toprule
    \multirow{2}[2]{*}{\textbf{Data}} & \multirow{2}[2]{*}{\textbf{Platform}} & \multicolumn{3}{c|}{\textbf{word2vec (skip)}} & \multicolumn{3}{c|}{\textbf{word2vec (cbow)}} & \multicolumn{3}{c|}{\textbf{doc2vec (skip)}} & \multicolumn{3}{c|}{\textbf{doc2vec (cbow)}} & \multicolumn{3}{c|}{\textbf{Asm2Vec}} & \multicolumn{3}{c}{\textbf{Ours}} \\
          &       & \textbf{P} & \textbf{R} & \textbf{F1} & \textbf{P} & \textbf{R} & \textbf{F1} & \textbf{P} & \textbf{R} & \textbf{F1} & \textbf{P} & \textbf{R} & \textbf{F1} & \textbf{P} & \textbf{R} & \textbf{F1} & \textbf{P} & \textbf{R} & \textbf{F1} \\
    \midrule
          & \textbf{ARM} & 0.44  & 0.48  & 0.46  & 0.43  & 0.49  & 0.46  & 0.57  & 0.71  & 0.63  & 0.63  & 0.72  & 0.68  & 0.62  & 0.76  & 0.68  & \textbf{0.87 } & \textbf{0.89 } & \textbf{0.88 } \\
          & \textbf{Aarch64} & 0.29  & 0.41  & 0.34  & 0.45  & 0.37  & 0.40  & 0.43  & 0.60  & 0.50  & 0.43  & 0.61  & 0.50  & 0.59  & 0.66  & 0.62  & \textbf{0.77 } & \textbf{0.83 } & \textbf{0.80 } \\
    \textbf{Vision} & \textbf{Android} & 0.42  & 0.50  & 0.46  & 0.41  & 0.51  & 0.46  & 0.57  & 0.71  & 0.63  & 0.61  & 0.71  & 0.66  & 0.64  & 0.65  & 0.64  & \textbf{0.84 } & \textbf{0.88 } & \textbf{0.86 } \\
          & \textbf{Intel} & 0.42  & 0.48  & 0.45  & 0.43  & 0.48  & 0.45  & 0.57  & 0.68  & 0.62  & 0.58  & 0.69  & 0.63  & 0.62  & 0.73  & 0.67  & \textbf{0.84 } & \textbf{0.87 } & \textbf{0.86 } \\
          & \textbf{mixed} & 0.39  & 0.47  & 0.42  & 0.43  & 0.46  & 0.44  & 0.53  & 0.68  & 0.60  & 0.56  & 0.68  & 0.62  & 0.62  & 0.70  & 0.65  & \textbf{0.83 } & \textbf{0.87 } & \textbf{0.85 } \\ \midrule
          & \textbf{ARM} & 0.41  & 0.41  & 0.41  & 0.38  & 0.36  & 0.37  & 0.51  & 0.44  & 0.48  & 0.47  & 0.43  & 0.45  & 0.58  & 0.59  & 0.59  & \textbf{0.76 } & \textbf{0.71 } & \textbf{0.73 } \\
          & \textbf{Aarch64} & 0.33  & 0.33  & 0.33  & 0.29  & 0.21  & 0.24  & 0.48  & 0.42  & 0.45  & 0.42  & 0.37  & 0.39  & 0.41  & 0.51  & 0.45  & \textbf{0.73 } & \textbf{0.65 } & \textbf{0.69 } \\
    \textbf{Textual} & \textbf{Android} & 0.51  & 0.37  & 0.43  & 0.46  & 0.47  & 0.46  & 0.42  & 0.45  & 0.43  & 0.48  & 0.43  & 0.45  & 0.57  & 0.61  & 0.59  & \textbf{0.67 } & \textbf{0.69 } & \textbf{0.68 } \\
          & \textbf{Intel} & 0.35  & 0.35  & 0.35  & 0.36  & 0.32  & 0.34  & 0.47  & 0.45  & 0.46  & 0.46  & 0.43  & 0.45  & 0.58  & 0.57  & 0.57  & \textbf{0.71 } & \textbf{0.71 } & \textbf{0.71 } \\
          & \textbf{mixed} & 0.40  & 0.36  & 0.38  & 0.37  & 0.34  & 0.35  & 0.47  & 0.44  & 0.45  & 0.46  & 0.42  & 0.44  & 0.54  & 0.57  & 0.55  & \textbf{0.72 } & \textbf{0.69 } & \textbf{0.70 } \\
    \bottomrule
    \end{tabular}%
}
\caption{Accuracy results in searching semantic similar functions}
  \label{tab:accuracy}%
\end{table*}%

\subsection{DNN Reconstruction}
\label{sec:rebuild}

\begin{algorithm}[hb!]
	\renewcommand{\algorithmicrequire}{\textbf{Input:}}
	\renewcommand{\algorithmicensure}{\textbf{Output:}}
	\caption{DNNs Reconstruction Algorithm}
	\label{alg:reconstruction}
	\begin{algorithmic}[1]
    \REQUIRE VicProg. \COMMENT{Victim AI program }
    \REQUIRE $\phi_{binary}$ \COMMENT{Semantic representation model}
    \STATE CG = ParseFile(VicProg) 
	   \STATE $N$ = CollectKernelNodes(CG)
        \WHILE{$N$ is not empty}
            \STATE $c$ = SelectNode(N) \COMMENT{select a node whose all precursors are computed}
            \STATE  $f =$ FindBinaryFunction($c$) 
            \STATE $k = \phi_{binary}(f)$ \COMMENT{Infer kernel type based on Eq. \ref{eq:cosine}}
            \STATE $c.type = k$  \COMMENT{Assign kernel to node $c$}
            \STATE Compute node $c$
            \STATE Remove $c$ from $N$
        \ENDWHILE
		
		\ENSURE  Return computed tensor of the output node in $CG$.
	\end{algorithmic}  
\end{algorithm}

\noindent After training a semantic representation model $\phi_{binary}$.
We propose a automatic algorithm to rebuild the victim DNNs.
As shown in Algorithm \ref{alg:reconstruction}, the inputs of our algorithm are the victim AI program and a trained semantic representation model.
We first parse the program to collect the graph files~($CG$) and collect all kernel nodes $N$ in $CG$ based on whether the node has an edge point.
While list $N$ is not empty, we select a node $c$ from N based on whether $c$'s all precursors have been inferred~(line 3).
For the selected node $c$, we find the corresponding binary function $f$~(line 4) and apply our semantic representation model to infer the kernel type $k$  of this function~(line 6).
After that, we compute the output of node $c$ and remove it from $N$.
We repeat this process iteratively until $N$ is empty.
Finally, we return the computed tensor of the output node in CG.
By doing so, we reconstruct the DNNs from the AI programs.

\vspace{-3mm}
\section{Evaluation}

We answer following questions through empirical evaluation.

\begin{itemize}
    \item \textbf{\textit{RQ1:}} \textit{How accurate of proposed representation model?}
    \item \textbf{\textit{RQ2:}} \textit{Can our algorithm 2 rebuild DNNs correctly?}
    \item \textbf{\textit{RQ3:}} \textit{Is our approach sensitive to hyper-parameters.}
    \item \textbf{\textit{RQ4:}} \textit{What's the contribution of each component?}
\end{itemize}

\subsection{Experimental Setup}

\begin{table}[tbp]
  \centering

  \resizebox{0.45\textwidth}{!}{
\begin{tabular}{r|l|cccc}
    \toprule
    \multicolumn{1}{l|}{\textbf{Type}} & \textbf{Platform} & \textbf{\# Train} & \textbf{\# Test} & \textbf{Instructions} & \textbf{Blocks} \\
    \midrule
          & \textbf{Android} & 22948 & 5737  & 194.75 ±  223.64 & 43.95 ± 47.19 \\
          & \textbf{Intel} & 22892 & 5724  & 141.52 ± 135.41 & 42.49 ± 45.61 \\
    \multicolumn{1}{l|}{\textbf{Vision}} & \textbf{ARM} & 28481 & 7121  & 208.19 ± 291.24 & 40.81 ± 44.70 \\
          & \textbf{Aarch64} & 5761  & 1441  & 197.68 ± 216.46 & 43.87 ± 47.24 \\
          & \textbf{Mix} & 80082 & 20023 & 195.61  ± 238.88 & 42.92 ± 46.41 \\
    \midrule
          & \textbf{Android} & 2647  & 662   & 131.48 ± 144.01 & 32.22 ± 38.23 \\
          & \textbf{Intel} & 808   & 207   & 97.01 ± 104.40 & 32.96 ± 38.058 \\
    \multicolumn{1}{l|}{\textbf{Textual}} & \textbf{ARM} & 2728  & 683   & 126.30 ± 142.88 & 21.20 ± 38.06 \\
          & \textbf{Aarch64} & 2618  & 655   & 135.67 ± 150.20 & 32.30 ± 38.51 \\
          & \textbf{Mix} & 8801  & 2207  & 127.94 ± 142.73 & 31.32 ± 37.66 \\
    \bottomrule
    \end{tabular}%
    }
    \caption{Data Statistics of Assembly Functions}
    \label{tab:data}%
\end{table}%

\paragraph{Datasets.}
We first  download the open source models from \textit{TorchVision} and \textit{HuggingFace} as our dataset.
Our model dataset includes vision models~(\eg VGG-11, \etc) and textual models~(\eg BertSentenceClassification, \etc).
We then use TVM, a popular deep learning compiler~\cite{tvm} to compile deep neural networks into AI programs.
We consider four different deployment platforms (\eg, Android, Intel, ARM, and Aarch64), and we set the compilation optimization level from 0 to 4.
For the executable files of the compiled programs, we use radare2 to disassemble the binary functions into assembly instructions.
We split 80\% of the data as our training dataset and the rest 20\% for testing.
The statistic of our dataset can be found in \tabref{tab:data}.

\paragraph{Comparison Baselines.}
We compare our proposed method with six unsupervised semantic representation methods: \texttt{word2vec}, \texttt{doc2vec}, and \texttt{asm2vec}.
For \texttt{word2vec}, we use the average vectors of all instructions in the assembly function to represent the semantic.
For \texttt{word2vec} and \texttt{doc2vec}, we consider both \texttt{Skip-Gram} and \texttt{CBOW} models.
For a fair comparison, for each method, we set the dimensions of the embedding as 100.

\subsection{(RQ1) Results for Binary Function Searching}

We follow existing work~\cite{alon2018code2seq,alon2019code2vec} to apply \textit{Semantic Precision}, \textit{Semantic Recall} and \textit{Semantic F-1} score as our metrics.
\tabref{tab:accuracy} summarizes the main results.
From the results, we could observe that: the proposed method significantly improves the binary function mapping accuracy across two datasets on all hardware platforms,
The observation demonstrates our embedding technique's success in enhancing the prediction capacity in the binary mapping problem.

\subsection{(RQ2) Reconstruct DNNs}

\begin{table}[tbp]
  \centering

  \resizebox{0.4\textwidth}{!}{
    \begin{tabular}{c|cc|c}
    \toprule
    Model & \multicolumn{1}{c}{\# of Layers} & \multicolumn{1}{c|}{\# of Parameters} & \multicolumn{1}{c}{Acc Loss} \\
    \midrule
    \textbf{LeNet-5} & 7     & 0.44M & 0.00 \\
    \textbf{ResNet-18} & 68    & 11.69M & 0.00 \\
    \textbf{ResNet-50} & 175   & 25.56M & 0.00 \\
    \textbf{MobileNet-V2} & 158   & 3.51M & 0.00 \\
    \textbf{VGG-11} & 29    & 132.86M & 0.00 \\
    \textbf{VGG-19} & 45    & 143,67M & 0.00 \\
    \textbf{WideResNet-50} & 175   & 68.88M & 0.00 \\
    \bottomrule
    \end{tabular}%
}
  \caption{Successfully reversed DNNs without accuracy loss}
  \label{tab:case}%
\end{table}%

In this section, we evaluate the effectiveness of our reconstruction algorithm (\ie Algorithm \ref{alg:reconstruction}).
Because a learning-based technique cannot guarantee soundness results, we choose DNNs with high kernel operator similarity scores as examples.
The selected model architectures are listed in \tabref{tab:case}.
We use the accuracy loss as our evaluation metric. Formally, we feed random inputs into the original and reconstructed models. Then, the accuracy loss is defined as the accuracy difference between the accuracy of the original model and the reconstructed model.
Table \ref{tab:case} lists some examples of our successful reversed DNN models.

\subsection{(RQ3) Parameter Sensitivity}

\begin{figure}[htb!]
    \centering
    \includegraphics[width=0.4\textwidth]{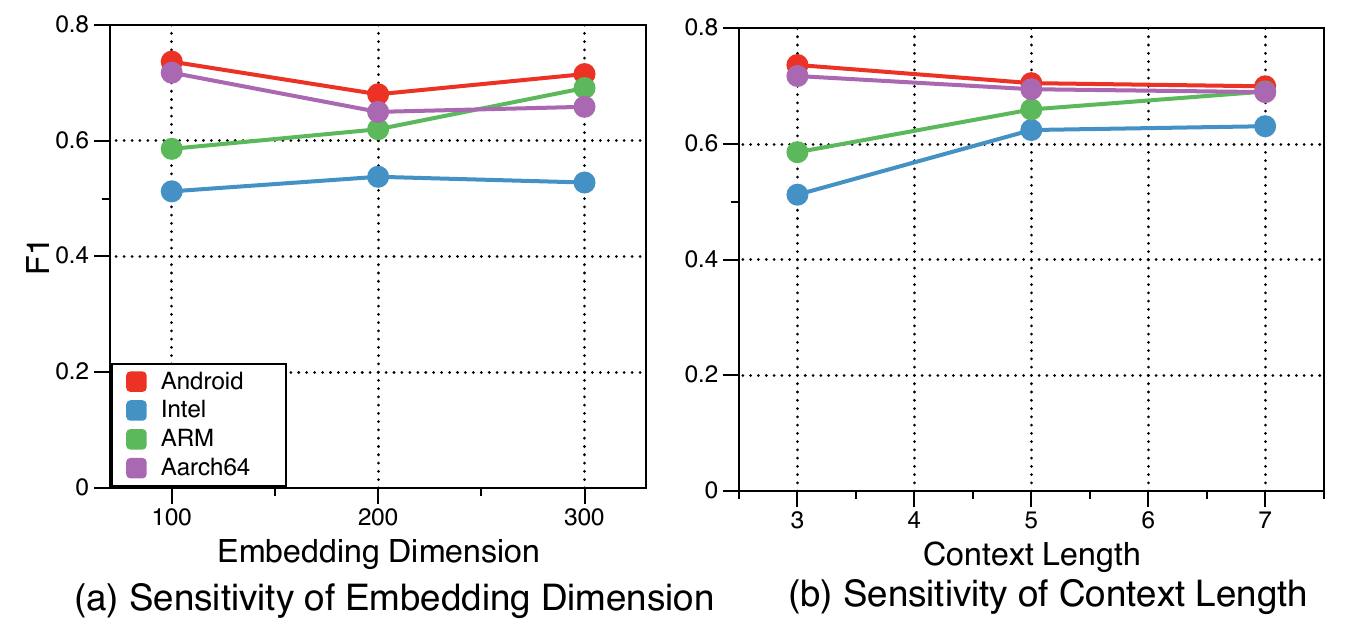}
    \caption{ Sensitivity of Representation Model }
    \label{fig:senstive}
\end{figure}

\noindent We perform the sensitivity analysis of our proposed techniques on the Textual dataset.
\textbf{(a) embedding dimension:} We first conduct experiments to understand how different embedding dimensions affect our technique. Specifically, we configure the embedding dimension as 100, 200, 300 and measure F-1 under different settings. \textbf{(b) context length:} We then configure the  context size as 3, 5, 7 and train different models. The results of different embedding configure are listed in \figref{fig:senstive}.  From the results, we observe that \tool can keep a stable F1 scores on different embedding dimension and context length settings, which indicates that \tool is not sensitive to embedding dimension and  context length.

\subsection{(RQ4) Ablation Studies}

\begin{table}[tbp]
  \centering

  \resizebox{0.33\textwidth}{!}{
    \begin{tabular}{l|cc|cc}
    \toprule
    \multirow{2}[2]{*}{\textbf{Platform}} & \multicolumn{2}{c|}{\textbf{Vision}} & \multicolumn{2}{c}{\textbf{Textual}} \\
          & \textbf{Our} & \textbf{Baseline} & \textbf{Ours} & \textbf{Baseline} \\
    \midrule
    \textbf{Android} & \textbf{0.21 } & 0.80  & \textbf{0.09 } & 0.71  \\
    \textbf{Intel} & \textbf{0.11 } & 0.60  & \textbf{0.62 } & 0.85  \\
    \textbf{ARM} & \textbf{0.12 } & 0.65  & \textbf{0.51 } & 0.98  \\
    \textbf{Aarch64} & \textbf{0.14 } & 0.75  & \textbf{0.08 } & 0.66  \\
    \textbf{Avg} & \textbf{0.14 } & 0.70  & \textbf{0.32 } & 0.80  \\
    \bottomrule
    \end{tabular}%
   }
  \caption{The OOV ratio results}
  \label{tab:oov}%
\end{table}%

\paragraph{Can fine-grained modeling eliminate Out of Vocabulary (OOV)?}
We first conduct experiments to explore whether our fine-grained instruction embedding can eliminate the issue of  OOV.
We model each assembly instruction as a ``word'' and
measure the OOV ratio, and then we measure the OOV ratio of our instruction embedding.
The OOV results are summarized in \tabref{tab:oov}. We observe that directly modeling each instruction as ``word'' will result in a high OOV ratio. 
This is because assembly instruction may contain the memory address, and the value of the address is had to predict.

\paragraph{The performance of each representation component?}
In this study, we compare our embedding method to solo text embedding and solo topological embedding to investigate the efficacy of embedding combinations.
From the results in Table \ref{tab:combine}, we observe that: (1) for most cases, 20 out of 30, combing text embedding and structure embedding can increase binary mapping accuracy. (2) structure embedding performs the poorest in all scenarios, implying that simply applying structure information is insufficient to infer function semantics.
CFG structures, on the other hand, are complementary to text, and combining them yields better results.

\section{Discussion \& Conclusion}

In this research, we present a technique for automatically reversing DNNs from deployed AI applications.
In particular, we propose a novel embedding technique for capturing the semantics of assembly functions. 
The experiments demonstrate that \tool is possible to correctly reverse  DNNs.

\section*{Acknowledgments}
This work was supported by NSF grant CCF-2146443.


\clearpage
\bibliographystyle{named}
\bibliography{ijcai22.bib}
\end{document}